\documentclass[smallextended]{svjour3}       

\usepackage[utf8]{inputenc} 
\usepackage[T1]{fontenc}    
\usepackage{hyperref}       
\usepackage{url}            
\usepackage{booktabs}       
\usepackage{amsfonts}       
\usepackage{nicefrac}       
\usepackage{microtype}      
\usepackage{lipsum}
\usepackage{graphicx}
\usepackage[numbers]{natbib}

\title{Attention-based Recurrent Neural Network for Urban Vehicle Trajectory Prediction}

\author{Seongjin Choi   \and	Jiwon Kim \and	Hwasoo Yeo }

\institute{Seongjin Choi \at
	Department of Civil and Environmental Engineering\\
	Korea Advanced Institute of Science and Technology\\
	\email{benchoi93@kaist.ac.kr}           
	\and
	Jiwon Kim \at
	School of Civil Engineering\\
	The University of Queensland\\
	\email{jiwon.kim@uq.edu.au}\\
	\and
	Hwasoo Yeo \at
	Department of Civil and Environmental Engineering\\
	Korea Advanced Institute of Science and Technology\\
	\email{hwasoo@kaist.ac.kr}           
}
\begin{document}
	\maketitle
	
	\begin{abstract}
		With the increasing deployment of diverse positioning devices and location-based services, a huge amount of spatial and temporal information has been collected and accumulated as trajectory data. Among many applications, trajectory-based location prediction is gaining increasing attention because of its potential to improve the performance of many applications in multiple domains. This research focuses on trajectory sequence prediction methods using trajectory data obtained from the vehicles in urban traffic network. As Recurrent Neural Network(RNN) model is previously proposed,  we propose an  improved method of Attention-based Recurrent Neural Network model(ARNN) for urban vehicle trajectory prediction. We introduce attention mechanism into urban vehicle trajectory prediction to explain the impact of network-level traffic state information. The model is evaluated using the Bluetooth data of private vehicles collected in Brisbane, Australia with 5 metrics which are widely used in the sequence modeling. The proposed ARNN model shows significant performance improvement compared to the existing RNN models considering not only the cells to be visited but also the alignment of the cells in sequence. 
	\end{abstract}

	\section{Introduction}
	Recently, with abundance of various location sensors and location-aware devices, a large amount of location data are collected in urban spaces. These collected data are studied in the form of so-called moving object trajectory which is a trace of moving object in geographical spaces represented by a sequence of chronologically ordered locations \cite{zheng_trajectory_2015}. Of particular interest are urban vehicle trajectory data that represent vehicle movements in urban traffic networks. Such urban vehicle trajectory data offer unprecedented opportunities to understand vehicle movement patterns in urban traffic networks by providing rich information on both aggregate flows (e.g., origin-destination matrix and cross-sectional traffic volume) and disaggregate travel behaviours including user-centric travel experiences (e.g., speed profile and travel time experienced by individual vehicles) as well as system-wide spatio-temporal mobility patterns (e.g., origin-destination pairs, routing information, and network traffic state) \cite{kim_spatial_2015}. Previous studies have used urban vehicle trajectory data to perform travel pattern analysis \cite{kim_spatial_2015, yildirimoglu_identification_2017} and develop real-world applications such as trajectory-based bus arrival prediction \cite{zimmerman_field_2011} and trajectory-based route recommendation system \cite{yuan_t-drive:_2013}.
	
	Among many applications of trajectory data mining \cite{mazimpaka_trajectory_2016}, this study focuses on trajectory-based location prediction problem. This problem concerns analyzing large amounts of trajectories of people and vehicles moving around a city to make predictions on their next locations \cite{noulas_mining_2012, gambs_next_2012, mathew_predicting_2012}, destinations \cite{krumm_predestination:_2006, krumm_predestination:_2007, horvitz_help_2012,xue_solving_2015, ziebart_navigate_2008}, or the occurrences of traffic related events such as traffic jams and incidents \cite{wang_prediction_2016}. In this study, we address the problem of predicting the sequence of next locations that the subject vehicle would visit, based on the information on the previous locations from the origin of the current trip and historical database representing the urban mobility patterns.
	
	Trajectory-based location prediction is gaining increasing attention from both academia and industry because of its potential to improve the performance of many applications in multiple domains. One example is Location-based Service (LBS). LBS uses location data of service users and provide user-specific information depending on the locations of service users. Typical examples of LBS include social event recommendation, location-based advertising, and location-based incident warning system. The location prediction can be applied to predictively provide information; for example, if a user’s next location is expected to be disastrous or congested, the service can inform the users to change route. Furthermore, when it is not possible to continue to provide service because the position of the user is lost due to sensor malfunctioning, predicting locations of the user can temporally replace the role of positioning system and continue the service \cite{monreale_wherenext:_2009, morzy_mining_2007}. Another example is the application on agent-based traffic simulators. Unlike traditional traffic simulators which consider traffic demand as input, an agent-based traffic simulator requires information on individual vehicle journey such as origin, destination, and travel routes \cite{martinez_agent-based_2015}. The result of vehicle location prediction can be used for real-time application of these agent-based traffic simulators. Vehicle location prediction can also be applied to inter-regional traffic demand forecasting. As the market of ride-sharing is continuously growing and Shared Autonomous Vehicles(SAV) are expected to appear on our road network in the near future, there is a strong need to predict inter-regional traffic demand so as to dispatch the proper number of SAV to areas of high demand. Location prediction model can be used to identify the demand hot spots by learning the mobility pattern of the users.
	
	In a previous work \cite{choi_network-wide_2018}, we proposed a Recurrent Neural Network (RNN) model to predict next locations in vehicle trajectories by adopting ideas from text generation model in natural language processing, where RNN has shown great success, and adapting them for use in our problem of location sequence prediction. The RNN model \cite{choi_network-wide_2018} considered the previously visited locations as the only input to predict the next location. Despite its simple structure, the model produced promising results. For instance, for more than 50\% of all the tested trajectory samples, our RNN model showed a high prediction accuracy in that the probability of correctly predicting the next location was greater than 0.7, whereas the referenced non-RNN model (used for performance comparison) showed the similar accuracy level only for less than 5\% of the tested samples \cite{choi_network-wide_2018}. To further improve the model performance, this study considers additional inputs that are likely to help predictions and proposes methodology that allows the incorporation of heterogeneous input sources into the existing RNN framework. A specific input that we consider in this study is the surrounding traffic conditions of a vehicle at the time when it starts its journey. Nowadays, drivers can easily observe the current traffic state in the urban traffic networks and plan their journey (choose their routes) by using various traffic information and routing services \cite{adler_investigating_2001, cabannes_impact_2018}. As a result, the location sequences (chosen routes) of individual vehicles are expected to be influenced by the network traffic conditions at the beginning of their journeys. Inspired by this idea, this study proposes an Attention-based RNN model, which embeds an attention interface to enable the RNN model to consider the current traffic state as an additional input to location prediction. A detailed explanation is in the Methodology section.

	\section{Methodology}
	\subsection{Representing Urban Vehicle Trajectories as Cell Sequences}
	Urban vehicle trajectory refers to a sequence of locations and times describing the path that a vehicle follows along its journey in urban traffic networks. Various sensors collect the location $(x,y)$ of vehicles and passage time($t$) to form the vehicle trajectory data. The data points in vehicle trajectories are continuous in space; that is, points are continuous-scaled coordinates of longitude and latitude. To learn movement patterns from a large amount of trajectory data, however it is necessary to define a finite set of representative locations that are common to all the trajectories that have the similar path. As such, the first step in building the trajectory prediction model is to discretize the vehicle trajectory data and convert each trajectory to a sequence of discretized locations. Based on the previous studies \cite{choi_network-wide_2018, kim_graph-based_2016, kim_trajectory_2017}, we partition the urban traffic network into smaller regions, or cells, so that continuous-scaled raw vehicle trajectory data are represented as discretized cell sequence data.
	
	Let $Tr=[(x_1,y_1 ),(x_2,y_2 ),…,(x_l,y_l )]$ represent a raw vehicle trajectory consisting l number of data points, where data point $(x_i,y_i)$ represents the longitude and latitude coordinate of the vehicle’s $i^{th}$ position. Using a large number of vehicle trajectory data allows the data points in all trajectories to combine and cluster in space based on the desired radius, denoted by $R$. Accordingly, the distance between the centroid of the point cluster and its farthest member point is approximately R for each spatial cluster. The centroid is the mean location of the data points within the cluster, and by using Voronoi tessellation method, the cell boundaries of the clusters (Voronoi polygons) are determined. 
	
	Given N cells in the network, a vehicle trajectory can be expressed as a cell sequence $[c_1,c_2,…,c_m ]$, where $c_j$ is the index of the $j^{th}$ visited cell $(1 \leq c_j \leq N)$ within trajectory $Tr$. Since each of the visited cells covers multiple trajectory data points, the length of the cell sequence ($m$) is always less than or equal to the length of original vehicle trajectory ($l$) (i.e. $m \leq l$). In addition to the cell sequence covering the original trajectory, two virtual spatial tokens $\#start$ and $\#end$ are added to the front and the back of the cell sequence. These virtual tokens are treated as virtual cells that do not exist in the actual network but only indicate the start and the end of the trip.
	
	The cell sequence is then separated into input vector $X$ and output label vector $Y$ for training, validating, and testing. Given the cell sequence containing $m+2$ cells including the start and the end tokens, input vector $X$ consists of first m+1 cells and output label vector $Y$ consists of $m+1$ elements starting from the second element ($c_1$). 

	\begin{equation}
		\begin{array}{lcl}
			&X&=[X_0,X_1,X_2,...,X_m ]\equiv [\#start,c_1,c_2,...,c_m ] \\[6pt]
			&Y&=[Y_0,Y_1,Y_2,...,Y_m ]\equiv [c_1,c_2,...,c_m,\#end]
		\end{array}
	\end{equation}
	
	\begin{figure}
		  \includegraphics[width=\textwidth]{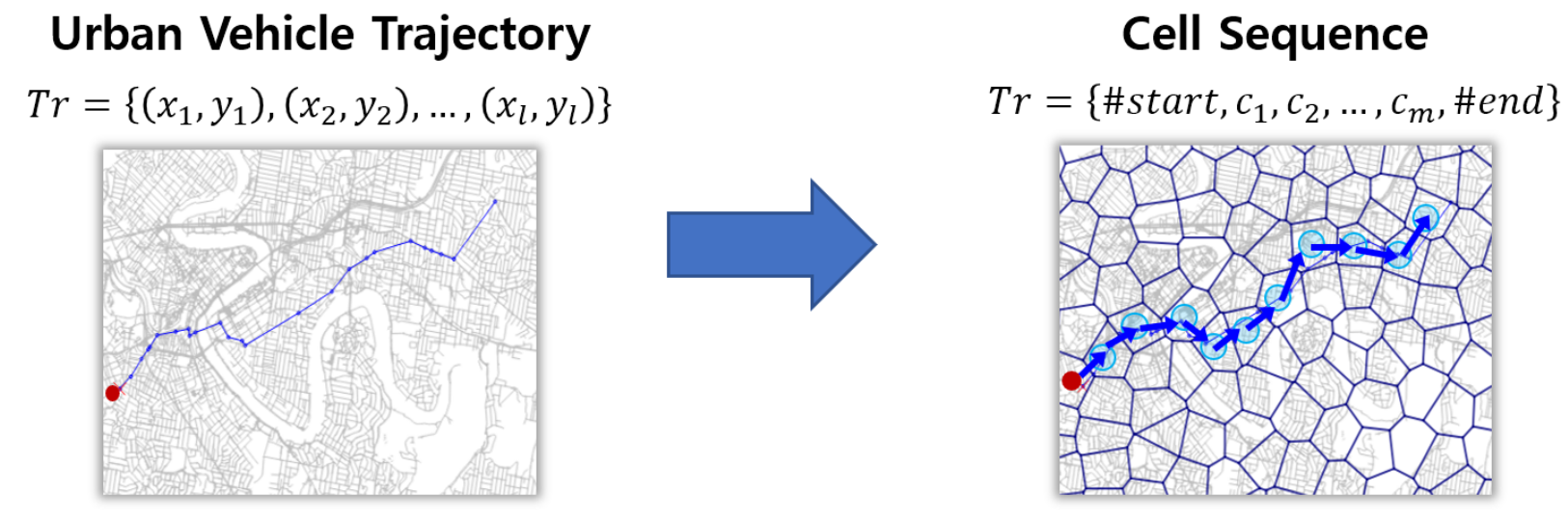}
		\caption{Representing urban vehicle trajectory as cell sequence}
		\label{figure1}
	\end{figure}
	
	\subsection{Cell Sequence Prediction using Recurrent Neural Network}
\begin{figure}
		  \includegraphics[width=0.75\textwidth]{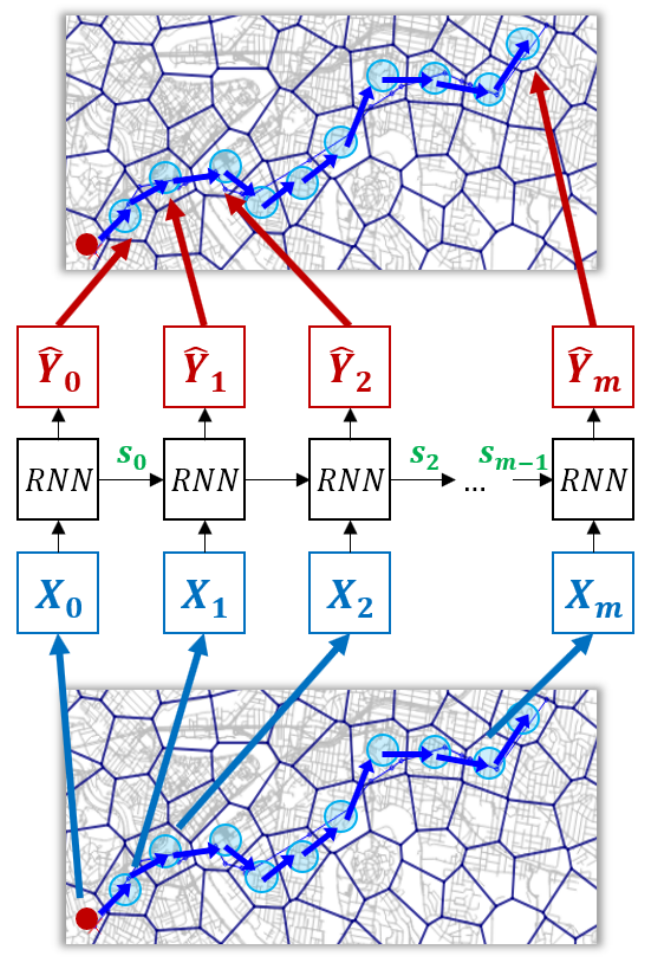}
		\caption{Structure of the basic Recurrent Neural Network model (RNN) for cell sequence prediction}
		\label{figure2}
	\end{figure}

	In our previous work \cite{choi_network-wide_2018}, a Recurrent Neural Network (RNN) model for the cell sequence prediction was developed and evaluated. This previous RNN model was designed to predict the future cell sequences based purely on the previously visited cell sequence. In the training step, the model calculates the probability of each cell being visited in the next step ($\hat{Y}_i$) based on input vector $X$. The model structure is shown in the Fig. \ref{figure2}. Then, the model calculates the cross-entropy loss ($L(Y,\hat{Y} )$) between the correct label ($Y$) and the predicted label probability based on the current parameter ($\hat{Y}$). A basic Long Short Term Memory (LSTM) unit \cite{hochreiter_long_1997} is used as the hidden unit in the RNN model, i.e., RNN units in Fig. \ref{figure2}. For parameter estimation, the Adam optimizer was used to update the model parameters \cite{kingma_adam:_2014}. A detailed explanation of the earlier model can be found in our previous paper \cite{choi_network-wide_2018}.

	\subsection{Incorporating Network Traffic State Data into Cell Sequence Prediction}
	The drivers can easily obtain the current traffic state in the urban traffic networks, and plan their journey by using various traffic information and routing services \cite{adler_investigating_2001, cabannes_impact_2018}. For example, between routes A and B, a driver is likely to choose route A when route B is congested and vice versa. As a result, the location sequences (chosen routes) of individual vehicles are expected to be influenced by the network traffic conditions at the beginning of their journeys. It is thus desirable to incorporate network-wide traffic state information and route choice behavior depending on the prevailing traffic state into the RNN-based cell sequence prediction model to increase the model’s prediction accuracy.
	
	Adding additional information to RNN models, which is network traffic state in our case, is not a straightforward task. RNN models are specialized to process sequential data considering temporal dependency across time or sequence steps. When input data are all in the form of sequence, adding another sequence input can be done through a straightforward extension as RNN model can have multiple input layers and multiple hidden features to incorporate multiple sequence inputs and combine them to calculate the output. However, when the additional input is non-sequential data, it cannot be directly represented as an input layer of the RNN model but rather should be processed outside the RNN model. The traffic state information we wish to add as an additional input to our RNN model is network-wide traffic density level at the beginning of the sequence, i.e., traffic information available at the origin ($X_0$) of a given trajectory), which is non-sequential data, making the problem more challenging. It may be possible to generate network traffic state data in a sequential form by feeding network traffic state at the time that a subject vehicle visits each cell in cell sequence. However, it requires a model that predicts the location and visiting time simultaneously and that is beyond the scope of our current study, which focuses on location prediction only.
	
\begin{figure}
		  \includegraphics[width=\textwidth]{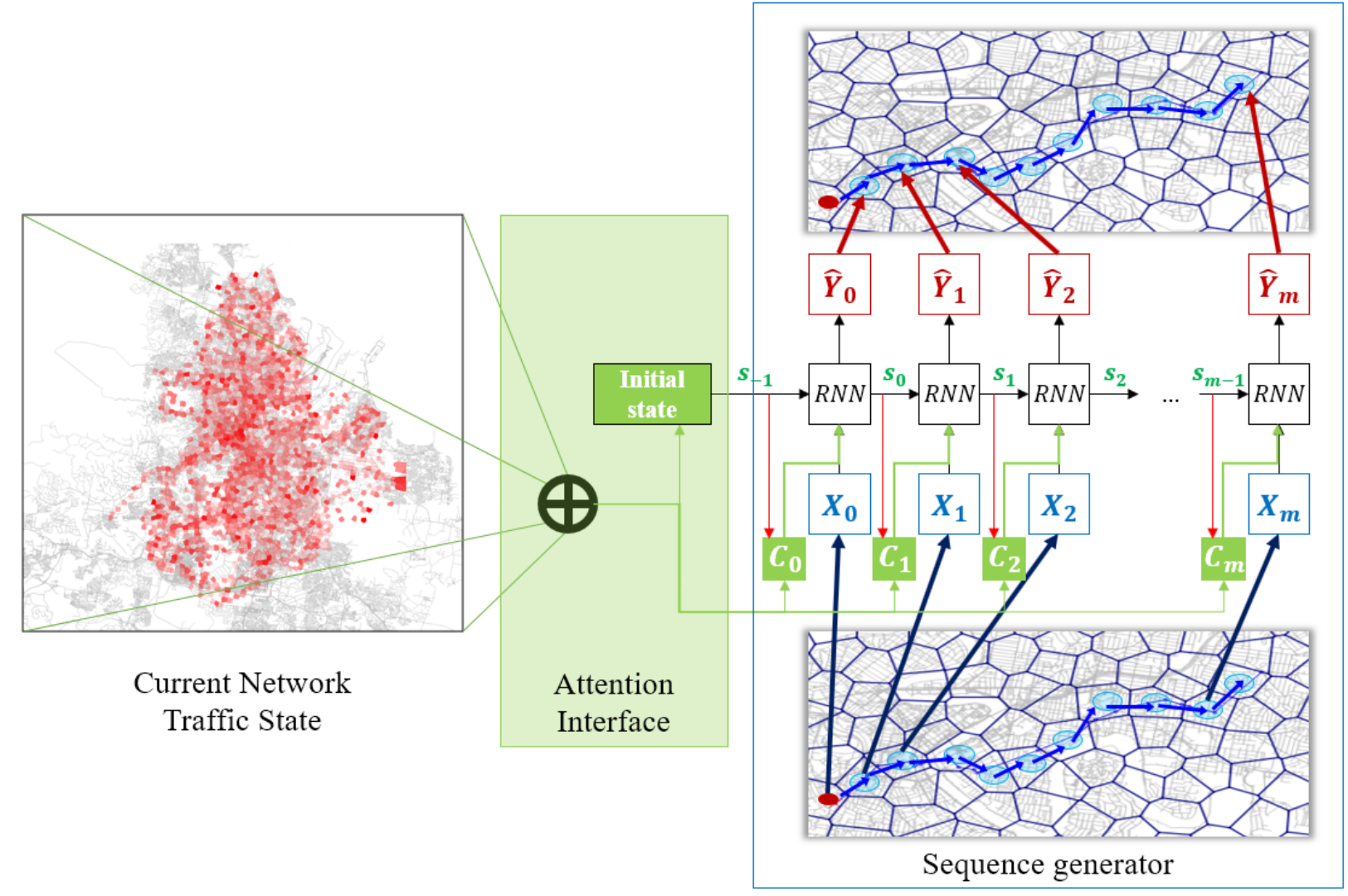}
		\caption{Structure of the proposed Attention-based Recurrent Neural Network model (ARNN) for cell sequence prediction}
		\label{figure3}
	\end{figure}

	One way to address this challenge is to introduce attention mechanism. The attention mechanism can be understood as an interface between external information processed outside the RNN model and sequential inputs processed inside the RNN model, as illustrated in Fig. \ref{figure3}. The attention mechanism in neural networks was first introduced to imitate the “attention mechanism” in human brain. When humans are asked to translate a sentence from one language to other language, humans try to think of words that match the alignment and meaning of word while also considering the global context of the sentence. Similarly, when humans are asked to write a sentence based on an image, humans not only concentrate on the important part of the image but also think of the global context of the image to write a sentence. Using the attention mechanism in neural networks has shown significant improvements in model performance in applications such as machine language translation \cite{vaswani_attention_2017} and video captioning \cite{xu_show_2015}.
	
	The attention mechanism allows the cell sequence prediction model or cell sequence generator to concentrate on certain part of the network traffic state input and use the information for cell sequence generation. There are mainly two tasks given to the attention mechanism: the first task is to set initial state for the RNN and the second one is to provide the network-wide traffic state information at each cell generation step. Usually, the initial state vector of RNN cell is set as zero vector since the simplest form of RNN does not consider additional information from other models or inputs. However, in the case of ARNN, there is an additional information of network traffic state. To consider this input in cell sequence generation, this information should be embedded into the model. Also, the attention mechanism allows the RNN to consider the traffic state in predicting the next location, or cell, at each step. The model is trained to calculate which information, or which region, to consider among the network traffic state data by calculating the context vector and attention weights.
	
	Fig. \ref{figure3} shows the structure of the Attention-based Recurrent Neural Network (ARNN) model for the cell sequence prediction. There are two types of input data in this model: the first is the network traffic state data and the second is the cell sequence representation of vehicle trajectory data. The model first processes the current network traffic state and calculates the initial state ($s_{-1}$) for the RNN unit. Then, the attention interface calculates the context vector ($C_i$) based on the previous state vector. The context vector ($C_i$) is used as input to the $i^{th}$ RNN unit as well as the corresponding input vector element ($X_i$) to update current state vector ($s_i$). The attention weight $\alpha_(i,j)$ is calculated based on the context vector and previous state vector ($ \alpha_(i,j)=f(C_i,s_(i-1)  )  $ ). The attention weight represents the probability to attend to $j^{th}$ cell at $i^{th}$ sequence. 9Therefore, the sum of $\alpha_(i,j)$ at each sequence is 1 ( $ \sum_{\forall j} \alpha_{(i,j)} = 1$ ). 
		
	The input cell sequence ($X$) is processed based on the word-embedding method to represent the hidden features of the cells. In the training step, input vector X is directly used as an input of each RNN unit in order to calculate the output vector ($\hat{Y}_i$). However, in the testing step, only the front n cell sequence elements are directly used. Afterward, since the output vector represents the probability of each cell being visited, we use a random sampling based on the multinomial distribution with probability $\hat{Y}_i$ to extract the next cell, also it is used as the next input vector element.
	
	A basic Long Short Term Memory (LSTM) cell \cite{hochreiter_long_1997} is used as RNN cell. And the model also uses the Adam optimizer to update the model parameters \cite{kingma_adam:_2014}.
	
	\section{Model Performance Evaluation}
	\subsection{Data}
	\subsubsection{Urban Vehicle Trajectory Data}
	The vehicle trajectory data used in this research were collected from the Bluetooth sensors in Brisbane, Australia, provided by Queensland Department of Transport and Main Roads (TMR) and Brisbane City Council (BCC). The Bluetooth sensors are installed in state-controlled roads and intersections located inside the Brisbane City, and they detect Bluetooth devices (e.g., in-vehicle navigation systems and mobile devices) passing the sensors and record their passage time. By connecting the data points containing the same identifier of the Bluetooth device (MAC ID), the vehicle trajectories of individual vehicles can be constructed. Each vehicle trajectory represents a time-ordered sequence of Bluetooth sensor locations that a subject vehicle passes. If the corresponding vehicle does not move for more than an hour, it is considered that the vehicle trip has terminated. For this case study, we used the vehicle trajectory data collected in March 2016. There are approximately 276,000 trajectories in one day, and a total of 8,556,767 vehicle trajectories were collected in March 2016. We randomly sampled 200,000 vehicle trajectories for the training dataset, 10,000 vehicle trajectories for the validation dataset (used in hyper-parameter searching), and 200,000 vehicle trajectories for the testing dataset.
	
	Brisbane urban traffic network is divided into “cells” to use the vehicle trajectory clustering and cell partitioning method proposed in the previous research \cite{kim_graph-based_2016, kim_trajectory_2017}. The desired radius of the cells is set to be 300m. Accordingly, a total of 5,712 cells are generated. Among them, 2,746 cells are considered to be active since the rest of the cells are not visited by any vehicles in the historical data of vehicle trajectories. The vehicle trajectory data are processed and transformed into cell sequence data.

	\subsubsection{Network Traffic State Data}
	There are several ways to represent the network traffic state such as density and average speed. In this study, vehicle accumulation, which is understood as the density of cells, is used to represent the network traffic state. The vehicle accumulation for a given cell is estimated by counting the number of vehicles that are present within the cell at a given instant point in time. We processed the vehicle trajectory data and calculated the vehicle accumulation of each cell at each minute. The vehicle accumulation data are normalized by dividing the vehicle accumulation by the historical maximum number of vehicle accumulation in each cell. 
	
	The vehicle accumulation data are used as the network traffic state input to the ARNN model. When the ARNN model is trained through each cell sequences, the model receives the vehicle accumulation data on a whole network from 10 minutes before the start time of the sequence. As a result, the shape of the input vehicle accumulation data is [N,10], where N is the number of cells in the study network.

	\begin{figure}
		  \includegraphics[width=\textwidth]{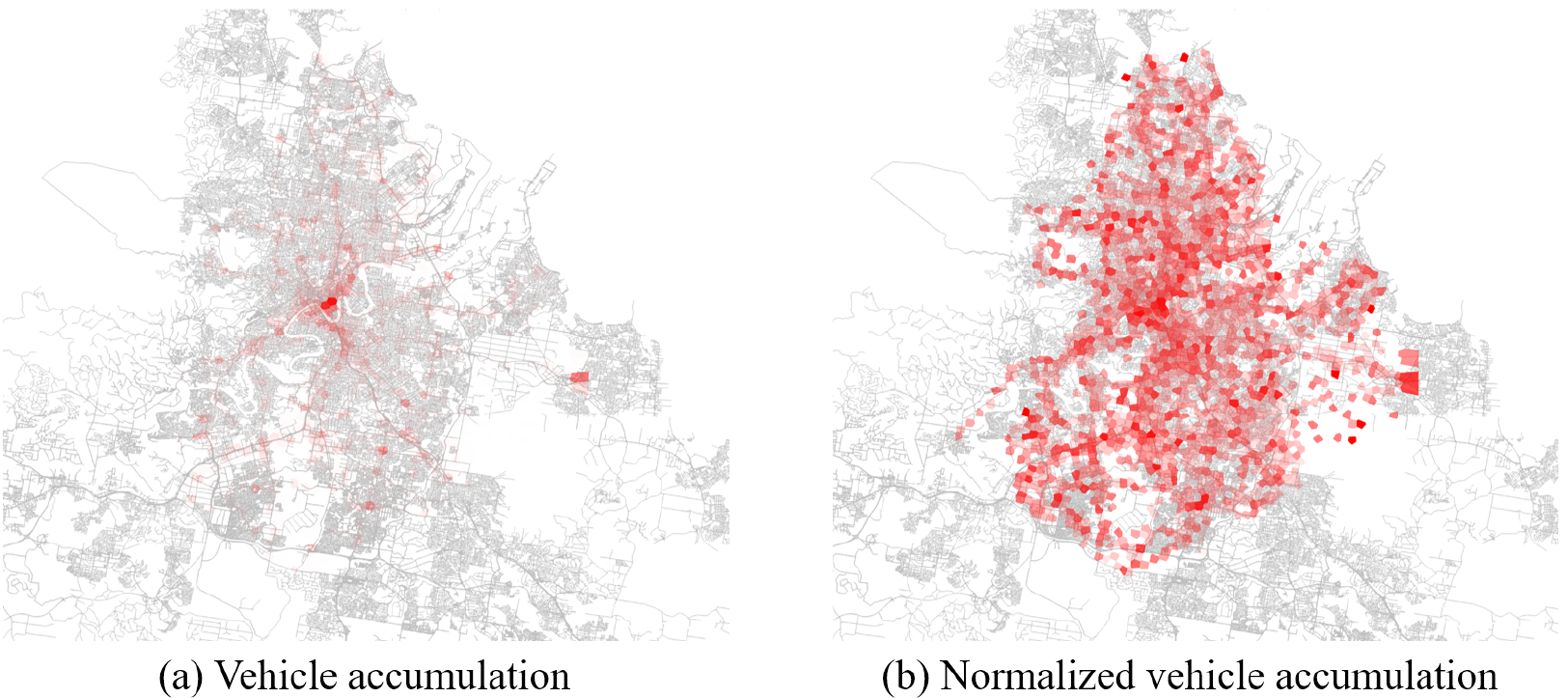}
		\caption{Spatial distribution of (a) vehicle accumulation, and (b) normalized vehicle accumulation at 12pm on March 1, 2016.}
		\label{figure4}
	\end{figure}

	\subsection{Hyperparameter and Model Training}
	For each model, we applied hyperparameter searching algorithm to ensure that each model is trained to achieve its maximum performance. The hyperparameter searching algorithm used in this study is from a Python package called “Scikit-optimize \cite{tim_head_scikit-optimize/scikit-optimize:_2018}.” The hyperparameter searching algorithm is based on Bayesian optimization using Gaussian Process (GP). This algorithm approximates the function by assuming that the function values follow a multivariate Gaussian. The covariance of the function values is given by a GP kernel between the parameters. Then a smart choice to choose the next parameter to evaluate can be made by the acquisition function over the Gaussian prior which is much quicker to evaluate. 
	
	There are three hyperparameters to search: learning rate, embedding layer dimension, and hidden layer dimension. The learning rate determines the updating step-size at each training step. If it is too large, the model is unlikely to converge. On the other hand, if it is too small, the speed of convergence is too slow, and the model is likely to fall into local minima. Therefore, finding an appropriate learning rate is crucial in learning neural networks. The embedding layer dimension is used to convert the cell sequence input which is treated as one-hot vector to a vector in the latent space. In other words, the embedding layer extracts feature of each cell input and represents it as a numeric vector. The hidden layer dimension determines the dimension of LSTM cells and cell decoding layer. LSTM cell is used to calculate the state vector ($s_i$) and cell decoding layer is used to calculate the cell-visiting probability ($\hat{Y}_i$) from the state vector ($s_i$). 
	
	The models are trained for 10 epochs for each hyperparameter set and the prediction accuracy is measured by applying the trained models to the validation dataset. The hyperparameters found are shown in Table \ref{table1}.

	\begin{table}[h]
		\caption{Hyperparameters}
		\begin{tabular*}{\hsize}{@{\extracolsep{\fill}}llll@{}}
			\toprule
			Model &  Learning rate &  Dimension of Embedding Layer & Dimension of Hidden Layer\\
			\midrule
			RNN   &  6.216234e-05 &  413 &854\\
			ARNN &  5.842804e-04 &  659 &574\\
			\bottomrule
		\end{tabular*}
		\label{table1}
	\end{table}

	\section{Result}
	\subsection{Score based Evaluation of Generated Cell Sequences}
	In this study, we use two widely used evaluation metrics in sequence modeling to evaluate the accuracy of generated cell sequences: BLEU score \cite{papineni_bleu:_2002}  and METEOR score \cite{banerjee_meteor:_2005}.
	In the previous study \cite{choi_network-wide_2018}, we used the complementary cumulative distribution function of the probability to measure how accurately the model predicts the next 1, 2, or 3 consecutive cells. While this measure is intuitive and easy to interpret, there is a drawback in this method in that it considers element-wide prediction accuracy and does not take the whole sequence  into account. The element-wide performance measures can be sensitive to small local mis-predictions and tend to underestimate the performance of the model. For example, when the original cell sequence is $[\#start,c_1,c_2,c_3,\#end]$ and a model is asked to predict the next cells based on the given cell sequence $[\#start,c_1 ]$, the prediction of $[c_2,c_4,c_3,\#end]$ will be considered as incorrect and performing poorly by our previous method because the model miss-predicted one cell $c_4$, even though the overall sequence is very similar to the original sequence. 
	As such, this study employs  $BLEU$ score and $METEOR$ score that consider the whole sequence, they are more robust and accurate as performance measure for sequence modeling. 
	
	\subsubsection{BLEU score}
	When a reference sequence is given, $BLEU$ uses three methods to evaluate the similarity between the reference sequence and the generated sequence. This metric is one of the most widely used metrics in natural language processing and sequence-to-sequence modeling. $BLEU$ scans through the sequence and check if the generated sequence contains identical chunks which are found in the reference sequence. Here, $BLEU$ uses a modified form of precision to compare a reference sequence and a candidate sequence by clipping. For the generated sequence, the number of each chunk is clipped to a maximum count ($m_{max}$) to avoid generating same chunks to get higher score.

	\begin{equation}
		\begin{array}{lcl}
			P_n=\frac{ \sum_{i \in C}   \min{(m_i,m_{i,max})}   }{w_t}
		\end{array}
	\end{equation}
	
	where $C$ is a set of cells (or chunks) in the generated sequence, $m_i$ is the number of the cell (or chunk) $i$ in the generated sequence, $m_{i, max}$ is the number of the cell (or chunk) $i$ in reference sequence, and $w_t$ is the total number of cells in candidate sequence. When $n$ is 1, the chunks represent the cells in the sequences. Otherwise, we consider $n$ consecutive cells as chunk and calculate the precision for each $n$-cell-unit.
	
	The $BLEU-n$ score represents the geometric mean of $P_i$'s with different $i$'s multiplied by a brevity penalty to prevent very short candidates from receiving too high score. 
	
	\begin{equation}
		\begin{array}{lcl}
			BLEU_n = min(1, \frac{L_{gen}}{L_{ref}}  ) \cdot (\prod_{i=1}^{n} P_i)^{\frac{1}{n}}
		\end{array}
	\end{equation}
	
	where $L_{gen}$ represents the length of generated sequence, $L_{ref}$ represents the length of reference sequence.

	\subsubsection{METEOR score}
	$METEOR$ \cite{banerjee_meteor:_2005} first creates an alignment between candidate cell sequence and reference cell sequence. The alignment is a set of mappings between the most similar cells. Every cell in the candidate sequence should be mapped to zero or one cell in the reference sequence. $METEOR$ chooses an alignment with the most mappings and the fewest crosses (fewer intersection between mappings). 
	
	To calculate $METEOR$ score, we first define precision $P$ and recall $R$.
	
	\begin{equation}
		\begin{array}{lcl}
			P=\frac{\min{(m,m_{max})}}{w_t}
		\end{array}
	\end{equation}

	\begin{equation}
		\begin{array}{lcl}
			R=\frac{\min{(m,m_{max})}}{w_r}
		\end{array}
	\end{equation}
	
	where \textit{m} refers to the number of single cells in candidate cell sequence which are also found in the reference cell sequence, $m_{max}$ refers to the sum of maximum number of each cell in reference cell sequence which are in candidate cell sequence, $w_t$ refers to the number of cells in candidate cell sequence, and $w_r$ refers to the number of cells in reference cell sequence. 
	
	Then, we calculate the weighted harmonic mean between precision and recall, where the ratio of the weights is 1:9.

	\begin{equation}
		\begin{array}{lcl}
			F_{mean}  = \frac{10}{\frac{1}{P} + \frac{9}{R}} = \frac{10PR}{R+9P}
		\end{array}
	\end{equation}

	To account for congruity with respect to a longer cell segment that appears both in reference and candidate cell sequences, we generate mappings based on the longer cell segment and use it to compute the penalty p. The more mappings there are, which are not adjacent in the reference and the candidate cell sequence, the higher the penalty will be. The penalty is calculated as follows:
			\begin{equation}
		\begin{array}{lcl}
			p=0.5 (\frac{c}{u_m})^3
			%
		\end{array}
	\end{equation}
		where \textit{c} is a set of single cells that are not adjacent in the candidate and reference sequence, and $u_m$ is the number of single cells that have been mapped. This penalty reduces $F_{mean}$ up to 50\% and calculate the $METEOR$ score (M).

	\begin{equation}
		\begin{array}{lcl}
			M=F_{mean}(1-p) 
		\end{array}
	\end{equation}
	
	\subsection{Score Result}

	For each sequence in the test dataset, the scores are calculated by the following procedure.
	
	Let $T_r$ be the subject cell sequence with length m, which is expressed as:
	
	\begin{equation}
		\begin{array}{lcl}
			T_r=[\#start,c_1,c_2,...,c_m,\#end]  
		\end{array}
	\end{equation}
	The subject cell sequence is divided into 2 parts: The sequence given $T_{r_{[1:g]}}$ and the sequence to be predicted $T_{r_{[g+1:m]}}$, where g is the number of cells given to the models (ARNN and RNN).
	\begin{equation}
		\begin{array}{lcl}
			&T_{r_{[1:g]}}=[\#start,c_1,c_2,...,c_g ]  
			\\
			&T_{r_{[g+1:m]}}=[c_{(g+1)},...,c_m,\#end]	
		\end{array}
	\end{equation}
	Each model predicts 100 candidate cell sequences based on $T_{r_{[1:g]}}$ producing a set of 100 $T_{r_{[1:m]}}$ sequences for each $T_{r_{[1:g]}}$. The generated candidate cell sequences are the cell sequences that have \#end token at the end, representing that the trip has terminated. These candidate cell sequences may not have the same length with the original cell sequence. The length can be longer or shorter depending on when the model predicts \#end token. The $T_{r_{[g+1:m]}}$ is used as reference cell sequence to calculate the score presented above. For each score ($BLEU_1$, $BLEU_2$, $BLEU_3$, $BLEU_4$, and $METEOR$), 100 score values are calculated based on the generated 100 candidate cell sequences. The average value of each score is used to represent the model performance of the corresponding cell sequence $T_r$. 
	
\begin{figure}
		  \includegraphics[width=\textwidth]{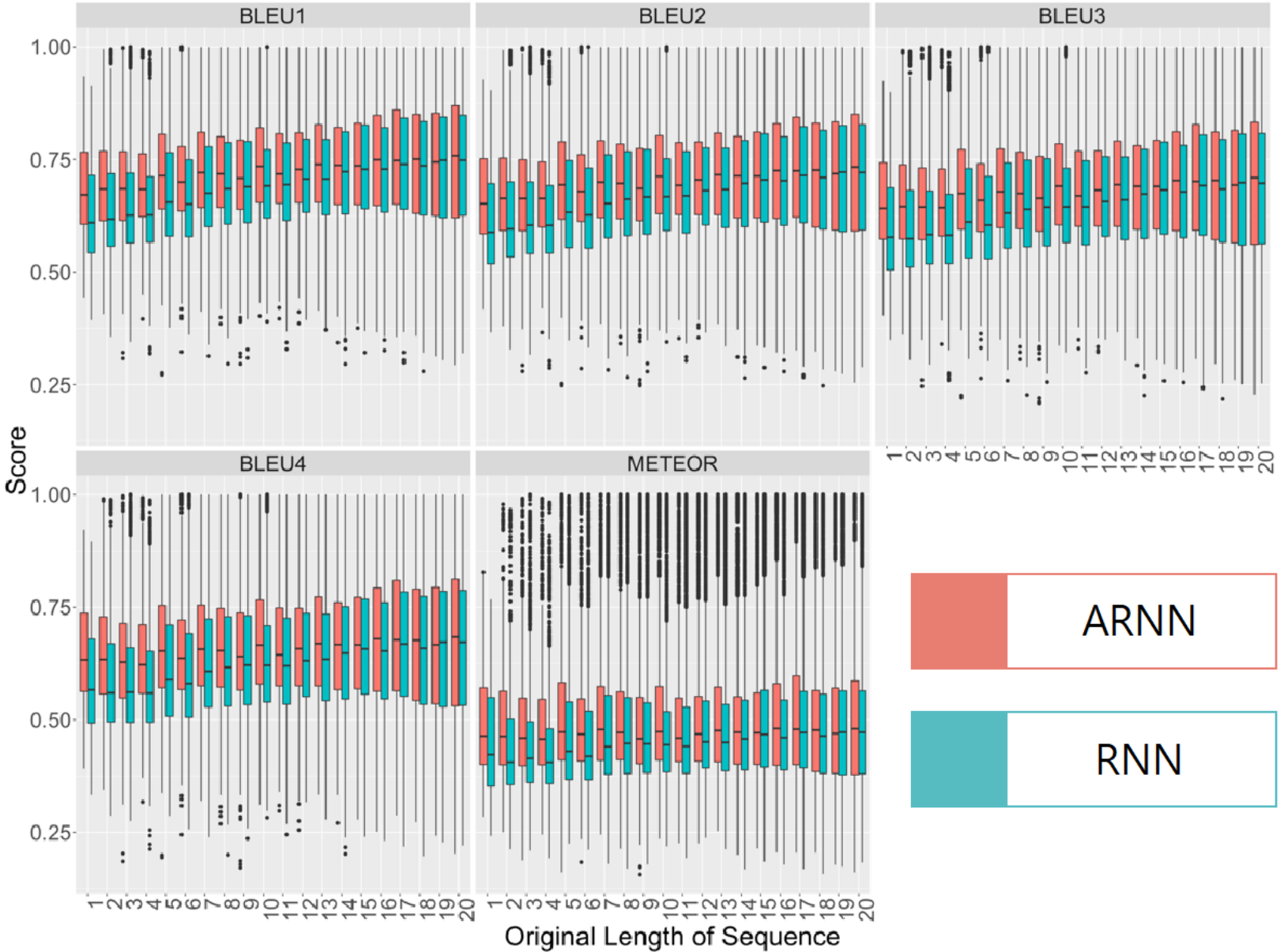}
		\caption{Box plot of models (ARNN, RNN) for each original length of sequence (m)}
		\label{figure5}
	\end{figure}

	10,000 cell sequences in test dataset is used to calculate the five scores.	Fig. \ref{figure5} shows the score result of each model. The x-axis represents the original length of the sequence, and the y-axis represents the value for each score metrics. The result of ARNN model (red color) shows better performance compared to the result of RNN model (blue color). 

	The result shows that ARNN model can predict short cell sequences more accurately up to 12\% and long cell sequences more accurately up to 5\%. ARNN model outperforms the RNN model in terms of both $BLEU$ score and $METEOR$ score. It is worth noting that the ARNN model had performance improvement in terms of $METEOR$ score as a high $METEOR$ score not only requires good prediction of visited cells but also accurate description of cell alignment (the visiting order of cells). The result thus confirm that the ARNN model using the attention mechanism achieves improvements in predicting the composition of cells in the sequence accurately as well as the alignment of the cells in the sequence.

\begin{figure}
		  \includegraphics[width=\textwidth]{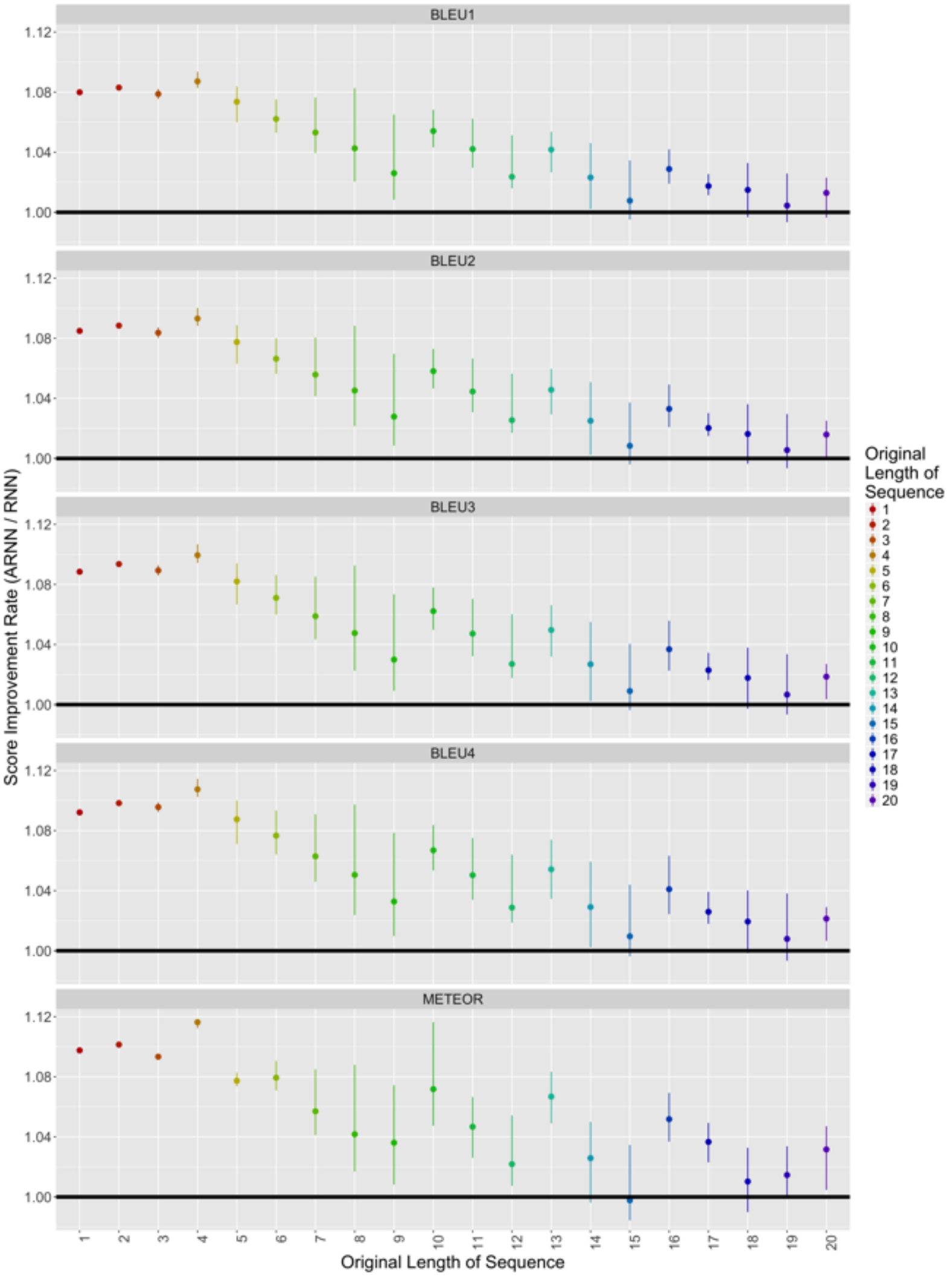}
		\caption{Score improvement rate for each original length of sequence (m). The points represent the average value, and the lines represent the range of the score improvement rate (from the minimum value to the maximum value)}
		\label{figure6}
	\end{figure}

	The performance gap between two models tends to decrease as the original length of the sequence increases. Fig. \ref{figure6} shows the result in terms of score improvement rate. The score improvement rate is defined as ratio of the performance score of ARNN model to the performance score of RNN model ($score_{ARNN}/score_{RNN}$). For each number of given cell sequence (g) and original length of cell sequence (m), this performance improvement rate is measured. And the Fig. \ref{figure6} shows the summarized result. Points in Fig. \ref{figure6} represents the average performance improvement rate for each original length of cell sequence (m), and the line represents the range of this value (from the minimum value to the maximum value).
	
	In Fig. \ref{figure6}, one interesting observation is that the performance improvement rate decreases and converges to 1 (the black lines) as the original length of the cell sequence increases. This can be because the input feature given to the ARNN model is the network traffic state at the beginning of each trajectory journey. This observation has an important implication for the influence of pre-trip information in route choice behaviors. The fact that the ARNN improves the prediction at the early stages of a journey implies that the pre-trip information indeed influences travellers’ route choice decisions and differentiates route choice patterns between different pre-trip traffic conditions. The fact that the effect of pre-trip information fades away at the later stages of the journey may be the indication of drivers’ reliance on en-route trip information instead of pre-trip information and thus indicates a need for incorporating such en-route information into the model to further improve the model performance.

	\clearpage
	
	\section{Conclusion and Future Studies}
	
	This research studies urban vehicle trajectory prediction, one of the applications of trajectory data mining. Based on the previous work \cite{choi_network-wide_2018}, in this study, we proposed a novel approach to incorporate network traffic state data into urban vehicle trajectory prediction model. Attention mechanism is used as an interface to connect the network traffic state input data to the vehicle trajectory predictor proposed in the previous work. ARNN model, which is Attention-based RNN model for cell sequence prediction, is compared with RNN model, which is RNN model for cell sequence prediction, in terms of conventional scoring methods in sequence prediction. The result shows that ARNN model outperformed RNN model.  It is effective to use attention mechanism to structurally connect the network traffic state input to RNN model to predict the vehicle’s future locations. Especially, it is promising that the ARNN model showed significant performance improvement in terms of METEOR which considers not only the cells to be visited but also the alignment of the cells in sequence. The performance improvement rates tend to decrease and converge to 1 as the original number of cell sequence increase. For the further improvement of the ARNN model, this problem should be studied to maintain the performance improvement rate at steady level.
	
	There are some limitations in this study, so further works should cover such topics. First of all, in this study, the network traffic state data were normalized by using the historical maximum value of each cell. This makes easy to represent the network traffic state, but this may lead to some problems that normalized data of the cells with very low traffic may be too sensitive to small number of vehicles and count it as heavy congestion. This makes the model overreact to these cells and makes the cell sequence prediction confused. For the further improvement of this study, different types of normalization methods should be tested. Second, although the performance measures ($BLEU$ and $METEOR$) are widely used in the fields studying sequence prediction such as natural language processing, the application of these metrics is new in the transportation domain. The interpretation and implication of these metrics in the context of traffic modeling should be further investigated and proposed. 
	

	\bibliographystyle{spbasic}      
	


\end{document}